\documentclass{article} 
\usepackage{iclr2026_conference,times}

\usepackage{amsmath,amsfonts,bm}

\def\eqref#1{equation~\ref{#1}}

\def\1{\bm{1}}

\DeclareMathAlphabet{\mathsfit}{\encodingdefault}{\sfdefault}{m}{sl}
\SetMathAlphabet{\mathsfit}{bold}{\encodingdefault}{\sfdefault}{bx}{n}

\usepackage{hyperref}
\usepackage{url}
\usepackage{graphicx}
\usepackage{graphicx}    
\usepackage{subcaption}
\usepackage{multirow}
\usepackage{booktabs} 
\usepackage[most]{tcolorbox}
\usepackage[ruled,vlined]{algorithm2e}

\newtcolorbox{takeaways}[1][]{
    enhanced,
    frame style={left color=blue!30, right color=blue!10},
    colframe=blue!50!black,
    colback=blue!5,
    arc=4pt,
    boxrule=1.5pt,
    title={\large\bfseries Takeaways}, 
    fonttitle=\sffamily,
    attach boxed title to top left={xshift=4mm, yshift=-2mm}, 
    boxed title style={
        colback=blue!50,
        colframe=blue!50!black,
        arc=3pt,
        left=4mm, 
        right=4mm, 
        boxsep=2pt, 
        before skip=0pt, 
        after skip=0pt 
    },
    #1
}

\title{ETTRL: Balancing Exploration and \\ Exploitation in LLM Test-Time Reinforcement \\ Learning  Via Entropy Mechanism}

\author{Jia Liu \\
Kuaishou Technology \\
liujiarik5@gmail.com \\
\And
ChangYi He \thanks{Work done during the internship at Kuaishou Technology.} \\
Beihang University \\
hechangyi@buaa.edu.cn \\
\And
YingQiao Lin \\
Kuaishou Technology \\
linyingqiao@kuaishou.com \\
\And
MingMin Yang \\
Kuaishou Technology \\
yangmingmin@kuaishou.com \\
\And
FeiYang Shen \footnotemark[1] \\
Northwestern Polytechnical University \\
shenfeiyang@mail.nwpu.edu.cn \\
\And
ShaoGuo Liu \thanks{Corresponding author} \\
Kuaishou Technology \\
sgliu2013@gmail.com \\
}

\iclrfinalcopy 
\begin{document}

\maketitle

\begin{abstract}
Recent advancements in Large Language Models (LLMs) have yielded to significant improvements in complex reasoning tasks such as mathematics and programming. 
However, these models remain heavily dependent on annotated data and exhibit limited adaptability in unsupervised scenarios.
To address these limitations, test-time reinforcement learning (TTRL) has been proposed, which enables self-optimization by leveraging model-generated pseudo-labels.
Despite its promise, TTRL faces several key challenges, including high inference costs due to parallel rollouts, and early-stage estimation bias that fosters overconfidence --- reducing output diversity and causing performance plateaus.
To address these challenges, we introduce an entropy-based mechanism to enhance the exploration–exploitation balance in test-time reinforcement learning through two strategies: Entropy-fork Tree Majority Rollout (ETMR) and Entropy-based Advantage Reshaping (EAR). 
Compared with the baseline, our approach enables Llama3.1-8B to achieve a 68\% relative improvement in Pass@1 metric on the AIME 2024 benchmark, while consuming only 60\% of the rollout tokens budget. 
This highlights our method's ability to effectively optimize the trade-off between inference efficiency, diversity, and estimation robustness, thereby advancing unsupervised reinforcement learning for open-domain reasoning tasks.
\end{abstract}

\section{Introduction}

Significant strides have recently been made in enhancing the reasoning capabilities of large language models (LLMs), particularly in expert-level domains such as mathematics and programming. These advancements are largely attributed to the convergence of two complementary paradigms:

\begin{enumerate}
  \item \textbf{Reinforcement Learning with Verifiable Rewards (RLVR)} --- exemplified by OpenAI-o1 \citep{jaech2024openai}, DeepSeek-R1 \citep{guo2025deepseek} and the Qwen3 family \citep{yang2025qwen3} --- is a paradigm that trains LLM policies using verifiable reward signals derived from final answers or intermediate reasoning steps. RLVR leverages either dense process-reward models (PRMs) \citep{lightman2023let} or sparse outcome-reward models (ORMs) to guide policy updates, enabling the model to refine its chain-of-thought (CoT) trajectories towards generating mathematically correct proofs or executable code \citep{wang2024math,cui2025entropy}.
  \item \textbf{Test-Time Scaling (TTS)} --- formalized by \citet{snell2025scaling} and \citet{liu2025can} --- is a paradigm that reallocates the FLOP budget from massive pre-training to \emph{inference-time} search. TTS strategies such as beam search, best-of-N sampling, and Monte-Carlo Tree Search (MCTS) allow a fixed model to expend additional compute at test time, often outperforming models 10–50$\times$ larger that rely solely on greedy generation. 
\end{enumerate}

Despite these successes, RLVR and TTS encounter several critical bottlenecks. RLVR depends on ground-truth datasets or at least verifiable outputs to generate reward signals, which restricts its applicability in fully unlabeled or distribution-shifted tasks where neither human annotations nor executable environments are available. Although TTS avoids additional training of the base model, it incurs substantial computational costs during inference and struggles to maintain output consistency.

To address these limitations, Test-Time Reinforcement Learning (TTRL) \citep{zuo2025ttrl} has recently emerged. During inference on an unseen prompt, TTRL repeatedly samples multiple candidate responses, derives a pseudo-label via majority voting, and performs on-the-fly policy gradient updates using these self-estimated rewards.

Consequently, TTRL provides a principled framework for lifelong, open-domain reasoning, enabling LLMs to autonomously refine their problem-solving strategies post-deployment. This capability allows models to solve novel challenging problems, without the need for labeled data or costly re-training.

However, TTRL currently suffers from two critical weaknesses:
\begin{enumerate}
  \item \textbf{High inference budget.} TTRL must perform tens to hundreds of rollouts to obtain a reliable pseudo-label. Consequently, mainstream parallel-estimation schemes incur prohibitive computational costs, and more challenging problems require even greater rollout budgets.
  \item \textbf{Early-stage estimation bias.} During the early iterations, the pseudo-label is often incorrect, yet the model may quickly overfit to it with greater advantage. This premature overconfidence drives the policy model into local optima and blocks further exploration.
\end{enumerate}

\begin{figure}[b]
  \begin{center}
  \includegraphics[width=1\textwidth]{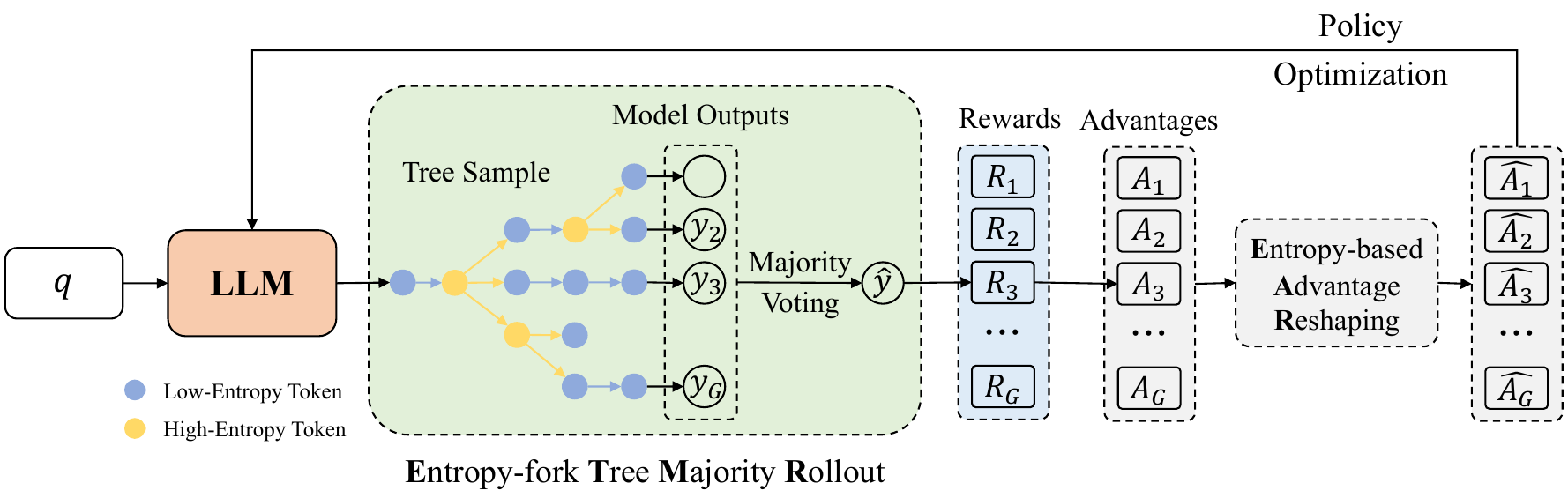}
  \end{center}
  \caption{The ETTRL framework employs an entropy-based majority voting mechanism to estimate pseudo-labels. During the advantage estimation phase, an entropy-based advantage shaping method is introduced, which balances exploration and exploitation in test-time reinforcement learning across two dimensions: the rollout process and reward signals.}
  \label{fig:ETTRL_example}
\end{figure}

To address the limitations of TTRL, we propose \textbf{Entropy-based Test-Time Reinforcement Learning (ETTRL)} framework. As illustrated in Figure \ref{fig:ETTRL_example}, ETTRL consists of two components:

\begin{enumerate}
    \item \textbf{Entropy-fork Tree Majority Rollout (ETMR)}: To tackle the high computational overhead and insufficient exploration of standard rollouts, we propose ETMR, a tree-structured rollout strategy that selectively branches only at the K tokens with the highest entropy (i.e., the ``fork points'' identified by \citet{hou2025treerl}). This mechanism generates a more diverse set of candidate responses with fewer tokens budget. On the AIME 2024 benchmark, ETMR enables Qwen2-1.5B to achieve a 5.24 percentage-point improvement in Pass@1 over the vanilla TTRL baseline, while halving the rollouts cost.
    \item \textbf{Entropy-based Advantage Reshaping (EAR)}: To mitigate the early estimation bias and sustain exploration, we introduce EAR. This method reshapes the advantage in Group Relative Policy Optimization (GRPO) \citep{shao2024deepseekmath} by incorporating a response-level relative entropy bonus into the calculation. The correction mitigates early-stage overestimation bias toward low-confidence rewards observed in vanilla GRPO, yielding an additional 3.0 percentage-point improvement in Pass@1 on AIME 2024.
\end{enumerate}

\section{Related Work}

Unlike verifiable reinforcement learning, test-time reinforcement learning faces two fundamental challenges: 1. \emph{Reward Estimation}: how to obtain reliable reward signals without explicit supervision. 2. \emph{Exploration-Exploitation Trade-off}: how to balance exploratory actions and reward exploitation during estimation.

\subsection{Unsupervised Reward Estimation}

Recent advances in large-scale reinforcement learning (RL) for reasoning tasks have centered on \emph{unsupervised reward estimation} --- the challenge of generating reliable reward signals without access to ground-truth labels, human feedback, or external verifiers. This research direction is driven by the prohibitive cost of expert annotation and the need for continual self-improvement in open-ended domains such as mathematics, code generation, and scientific reasoning. Below, we survey two dominant paradigms that have emerged: (1) \emph{entropy minimization} as an intrinsic reward; and (2) \emph{consensus-based} reward estimation via test-time scaling.

\paragraph{Entropy Minimization as Intrinsic Reward.}
The hypothesis that a model’s response confidence can serve as a proxy for correctness underpins a growing body of work in unsupervised RL. \citet{prabhudesai2025maximizing} introduced \textbf{RENT}, which uses the negative token-level response entropy of a language model as a dense reward. Experiments on GSM8K, MATH-500, AMC, AIME, and GPQA demonstrate consistent improvements across multiple model families (Qwen, Mistral, Llama) without any labeled data. \citet{agarwal2025unreasonable} extended this idea with three complementary techniques: (i) \textbf{EM-FT} --- direct fine-tuning by minimizing token-level entropy on self-sampled outputs; (ii) \textbf{EM-RL} --- policy-gradient RL using negative entropy as the sole reward; and (iii) \textbf{EM-INF} --- inference-time logit adjustment to reduce entropy without parameter updates. Notably, EM-RL matches or even surpasses the label-supervised baselines such as GRPO \citep{shao2024deepseekmath} and RLOO \citep{ahmadian2024back}, while EM-INF allows Qwen-32B to outperform GPT-4o on the challenging SciCode benchmark \citep{tian2024scicode}. These results corroborate earlier findings in unsupervised RL \citep{grandvalet2004semi} and domain adaptation \citep{wang2020tent}.

Despite these empirical successes, unsupervised reward estimation is still constrained by (i) the inductive biases of the base model \citep{agarwal2025unreasonable}, (ii) the alignment between confidence and correctness \citep{prabhudesai2025maximizing}, and (iii) the complexity of the target task domain \citep{zuo2025ttrl}.

\paragraph{Consensus-Based Reward Estimation via Test-Time Scaling.}
A parallel line of work leverages \emph{majority voting} or \emph{self-consistency} \citep{wang2022self} to generate pseudo-labels for RL. \citet{zuo2025ttrl} formalized this approach as \textbf{TTRL}, which optimizes the policy model on unlabeled test data using rewards derived from majority-voted answers. TTRL improves Qwen-2.5-Math-7B by 211\% on AIME 2024 and approaches the performance of supervised RL trained directly on ground-truth labels. The key insight is that, even when the majority answer is incorrect, reward accuracy can remain high due to the \emph{``lucky hit''} phenomenon --- incorrect predictions that disagree with the (wrong) consensus still receive the correct negative reward. This robustness to label noise aligns with theoretical analyses showing that RL can tolerate high error rates in reward models \citep{razin2025makes}. \citet{shao2025spurious} further demonstrated that even random rewards can yield non-trivial improvements under certain conditions, highlighting the importance of reward signal density over precision.

\subsection{Entropy Mechanism in Reinforcement Learning}

The role of entropy in reinforcement learning has been extensively studied across three complementary dimensions: (1) as a regularizer for balancing exploration and exploitation; (2) as a predictive indicator for scaling laws and performance ceilings; and (3) as a controllable variable that can be shaped to facilitate policy model optimization. We position our work within this landscape and highlight key advances that motivate our covariance-based entropy control framework.

\paragraph{Entropy as Exploration Signal}
Entropy has long been recognized as a principled metric for quantifying uncertainty and guiding exploration in reinforcement learning \citep{ziebart2008maximum,haarnoja2018soft}. In the context of large LLMs, recent studies have shown that policy entropy undergoes a predictable collapse during training, wherein rapid entropy decay correlates with early performance gains but eventually results in exploration stagnation \citep{cui2025entropy}. This phenomenon underscores the intrinsic tension between exploitation and exploration in policy optimization.

\paragraph{Entropy-Regularized Policy Optimization}
Traditional approaches to mitigating entropy collapse typically incorporate entropy regularization, in which an entropy bonus is added to the objective function \citep{schulman2017proximal,haarnoja2018soft}. However, these methods often require meticulous tuning of regularization coefficients and may destabilize training when applied directly to LLMs \citep{cui2025entropy}. Empirical evidence further indicates that entropy loss can either be ineffective or induce entropy explosion, thereby underscoring the necessity for more sophisticated entropy control mechanisms \citep{cui2025entropy}.

\paragraph{Entropy for Advantage Shaping}
Beyond direct regularization, entropy can also function as a signal for shaping policy advantages. \citet{cheng2025reasoning} demonstrated that high-entropy tokens are correlated with exploratory reasoning behaviors, such as pivotal logical connectors and self-reflection. It proposed an entropy-augmented advantage term that encourages longer reasoning chains without disrupting the original policy gradient flow. This method achieves superior performance on challenging benchmarks like AIME and AMC while maintaining computational efficiency.


\begin{takeaways}
    \begin{enumerate}
        \item Entropy minimization and consistency estimation constitute the primary methodologies for reward signal estimation in test-time reinforcement learning, representing the paradigms of soft and hard estimation, respectively. However, from a mechanistic standpoint, these approaches have not yet achieved an effective balance between exploitation and exploration.
        \item  Entropy reflects both epistemic uncertainty and exploratory potential, and incorporating entropy into either the reward or the advantage function can stabilize training and improve generalization. Building on these insights, we propose a lightweight entropy-shaping reward mechanism specifically designed for reasoning LLMs.
    \end{enumerate}
\end{takeaways}

\section{Methodology}

\subsection{Preliminaries}

\paragraph{Group Relative Policy Optimization (GRPO)}\label{subsec:grpo-detail}
Group Relative Policy Optimization \citep{shao2024deepseekmath} is an on-policy, advantage-based algorithm that fine-tunes LLMs without requiring an additional value model.  Below we give the full derivation, followed by its practical implementation.
Let $\pi_\theta$ denote the current policy and $\pi_{\text{old}}$ denote the behavioral policy used to collect the mini-batch.  
For each prompt $q$ and ground-truth answer $a$, GRPO samples $G$ complete responses $\{o_i\}_{i=1}^G\sim\pi_{\text{old}}(\cdot\mid q)$.  
For example, verifiable math problems are scored with a binary outcome reward for each response:

\begin{equation}
R_i=\mathbb{I}\bigl[\text{extract\_answer}(o_i)=a\bigr]\in\{0,1\}.
\label{eq:binary-reward}
\end{equation}

Then the \emph{group-relative} advantage for every token $t$ in response $o_i$ is:

\begin{equation}
\hat A_{i,t}=\frac{R_i-\mu}{\sigma},\quad
\mu=\frac{1}{G}\sum_{j=1}^{G}R_j,\quad
\sigma=\sqrt{\frac{1}{G}\sum_{j=1}^{G}(R_j-\mu)^2}.
\label{eq:advantage}
\end{equation}

The final surrogate loss is a per-token clipped objective:

\begin{equation}
\mathcal{L}_{\text{GRPO}}(\theta)=
-\mathbb{E}_{q,a,\{o_i\}}\!\Bigg[
\frac{1}{G}\sum_{i=1}^{G}\frac{1}{|o_i|}\sum_{t=1}^{|o_i|}
\min\Bigl(
r_{i,t}(\theta)\hat A_{i,t},
\operatorname{clip}\bigl(r_{i,t}(\theta),1-\epsilon,1+\epsilon\bigr)\hat A_{i,t}
\Bigr)
\Bigg].
\label{eq:grpo-full}
\end{equation}

where $r_{i,t}(\theta)=\pi_\theta(o_{i,t}\mid q,o_{i,<t})\big/\pi_{\text{old}}(o_{i,t}\mid q,o_{i,<t})$ is the importance weight and $\epsilon$ (typically $0.2$) controls the trust-region size. No KL penalty is used in the canonical GRPO formulation.

\paragraph{Entropy Measures}

Entropy measures the uncertainty of a language model at both micro (token) and macro (response) levels. 
At the token level, for a given context prefix $c=(q,o_{<t})$, the policy $\pi_\theta$ defines a categorical distribution over the vocabulary $\mathcal{V}$.  The Shannon entropy of the next-token distribution is: 

\begin{equation}
H_t=H\bigl[\pi_\theta(\cdot\mid c)\bigr]=-\sum_{v\in\mathcal{V}}\pi_\theta(v\mid c)\log\pi_\theta(v\mid c).
\label{eq:token-entropy-detailed}
\end{equation}

At the response level, let $o=(o_1,\dots,o_T)$ denote a complete response. The response-level entropy aggregates the token-level entropies while accounting for possible length variation:

\begin{equation}
H_{\text{resp}}(o)=\frac{1}{T}\sum_{t=1}^{T}H_t.
\label{eq:response-entropy-detailed}
\end{equation}

This metric has been shown to correlate with reasoning confidence \citep{wang2025beyond,agarwal2025unreasonable}.

\paragraph{Unsupervised Reward Estimation}\label{subsec:unsupervised-rewards}
When ground-truth labels are unavailable, we rely on \emph{intrinsic} or \emph{consensus-based} reward functions. Below we detail two representative approaches. (i)Minimum-entropy reward: RENT~\cite{prabhudesai2025maximizing} and EM-RL~\cite{agarwal2025unreasonable} replace the external verifier reward with the negative response entropy:

\begin{equation}
R_{\text{ME}}(o)=-\beta H_{\text{resp}}(o),
\label{eq:rent-reward-detailed}
\end{equation}

where $\beta>0$ is a tunable coefficient. Maximizing this reward discourages uncertain generations, implicitly guiding the model toward more confident --- and empirically more accurate reasoning paths without requiring any labeled data.
(ii) Test-Time Reinforcement Learning: TTRL \citep{zuo2025ttrl} performs RL on unlabeled test data by estimating rewards via \emph{majority voting}. The pipeline is as follows:
For a given prompt $q$, sample $N$ responses $\{o_i\}_{i=1}^N$ from the current policy. Extract the answer $y_i=\text{extract\_answer}(o_i)$ and compute the empirical majority label $y^\star$ over the discrete answer space $\mathcal{Y}$.
Each response is then assigned a binary reward according to:

\begin{equation}
R_{\text{TTRL}}(o_i)=\mathbb{I} \bigl[\text{extract\_answer}(o_i)=y^\star\bigr].
\label{eq:ttrl-reward-detailed}
\end{equation}

\subsection{ETMR: the exploration and exploitation of estimating pseudo-label}

Unsupervised reinforcement learning through consistency reward estimation has been successfully applied to reasoning tasks such as mathematics. However, this approach suffers from a notable limitation: during the estimation stage, a substantial token budget is required to obtain reliable pseudo-labels. For complex tasks which often require more than 64 rollouts to achieve reliable results, this demand is particularly costly, whereas supervised reinforcement learning typically requires fewer rollouts. We observe significant character-level repetition in the vocabulary generated by rollouts. Many rollouts contain substantial redundant tokens, which waste the valuable token budget allocated for testing and learning, thereby reducing overall training efficiency.

To address this, we explore methods for reusing duplicate tokens without compromising estimation accuracy. Recent research \cite{wang2025beyond} indicates that output diversity in reasoning is primarily influenced by high-entropy tokens --- typically conjunctions or transitional elements (e.g., ``but'', ``however''). In contrast, low-entropy tokens have minimal impact on final outcomes, particularly in verifiable reasoning tasks.

Building on this insight, we adapt the tree rollout methodology from TreeRL \citep{hou2025treerl}, which enables the reuse of low-entropy tokens during rollouts. Unlike traditional approaches that rely on explicit sentence-level segmentation, TreeRL employs token-based decision steps --- referred to as token steps --- to implicitly model the entire decision-making process. High-entropy tokens correspond to critical branching points that significantly influence reasoning quality, whereas low-entropy tokens can be efficiently reused. For high-entropy tokens, we select the top-K candidates to generate multiple sampling branches, thereby enabling fork-based exploration of diverse reasoning paths.

In our approach, all branches proceed to leaf nodes, ultimately generating complete responses. These responses are aggregated into candidate answers, and the final output is determined via a majority voting strategy. We refer to this approach as \textbf{Entropy-Fork Tree Majority Rollout}. By branching sampling trajectories at high-entropy tokens, this approach achieves greater sampling diversity with a lower token budget compared to conventional fully parallel sampling. The pseudocode for this process is provided below:

\begin{algorithm}[H]
\caption{Entropy-fork Tree Majority Rollout (ETMR)}\label{alg:treerl}
\KwIn{Prompt $x$, Policy $\pi_\theta$, Number of Trees $M$, Forking Points $N$, Branches $B$}
\KwOut{ $T$}
\For{$i \gets 1$ \KwTo $M$}{
    $Y^{(i)} \leftarrow \{y_i \sim \pi_\theta (\cdot | x)\}$ \\
    $T_i \leftarrow \{Y^{(i)}\}$ 
}
\ForEach{$T_i$}{
    $H(y_t) \leftarrow -\log \pi_\theta (y_t | x, y_{<t})$, $\forall t \in T_i$ \\
    $B_{i,l} \leftarrow \text{Top-}N H(\cdot | x) \{(t, H(y_t | x, y_{<t})) | t \in T_i\}$ \\
    \ForEach{selected forking point $(t, \cdot) \in B_{i,l}$}{
            $Y^{(i,l)}_{\text{new}} \sim \pi_\theta (\cdot | x, y_{<t})$ \\
            $T_i \leftarrow T_i \cup Y^{(i,l)}_{\text{new}}$, $j \in \{1, \cdots, T\}$
}
}
\end{algorithm}

In the process of ETMR, three key parameters $M$, $N$, and $B$ jointly determine the total number of rollout leaves. As defined, the final rollout count $R_{\text{tree}}$ is expressed in Equation \ref{eq:rollout-count}:

\begin{equation}
\label{eq:rollout-count}
R_{\text{tree}} = M(1+B*N)
\end{equation}

\begin{figure}[htbp]
  \centering
  \includegraphics[width=0.6\textwidth]{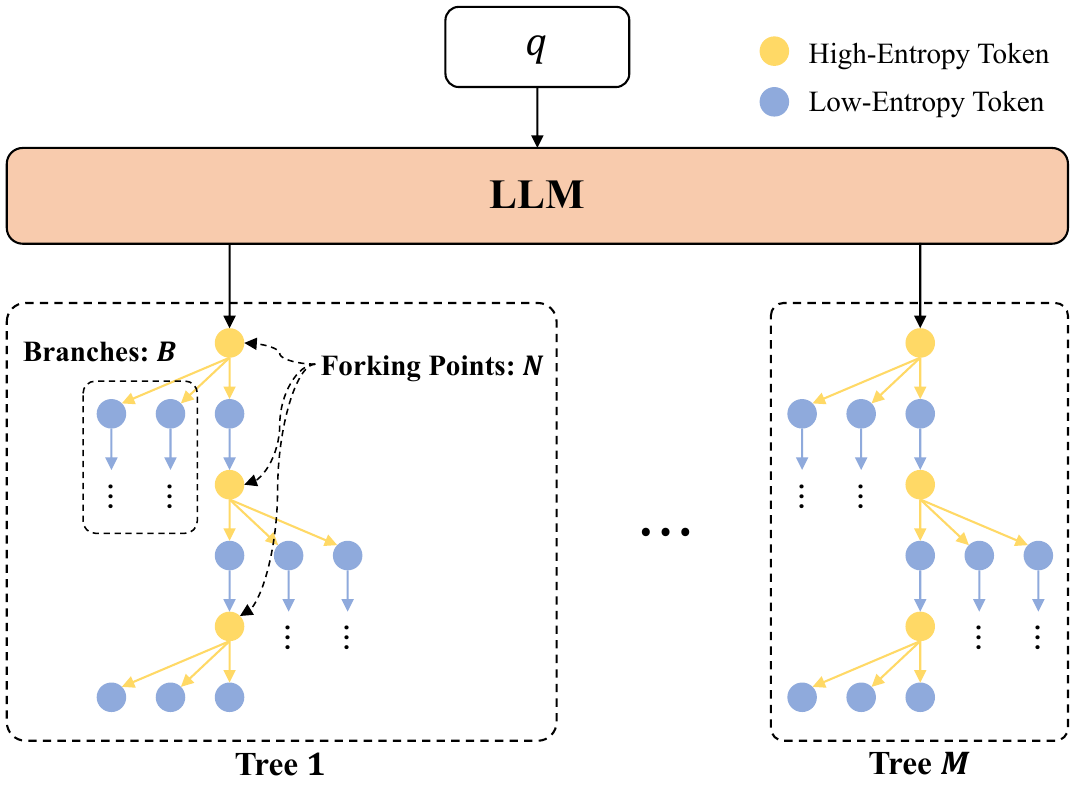}
  \caption{Entropy-fork Tree Sample (Forking Points $N = 3$, Branches $B = 2$)}
  \label{fig:MBN}
\end{figure}

Due to the positional uncertainty of the entropy fork points, the early forks result in a lower token reuse rate, whereas the later forks significantly enhance token reuse. To mathematically characterize the efficiency gains of ETMR algorithm, we assume that the entropy-fork points are uniformly distributed across the entire sampling process (as illustrated in Figure \ref{fig:MBN}). Consequently, the token consumption for a single tree-based rollout $T_{\text{tree}}$ can be modeled as an arithmetic sequence,as expressed in Equation \ref{eq:rollout-cost}. Here, $Len$ denotes the average response length and the sum of the arithmetic sequence depends solely on the number of entropy-fork points.

\begin{equation}
\label{eq:rollout-cost}
T_{\text{tree}} = Len*(1+B*\sum_{k=1}^N  k/(N+1))) \quad \text{where} \sum_{k=1}^N  k/(N+1)= N/2
\end{equation}

We further define the token consumption ratio per rollout $TR_{\text{tree}}$ in Equation \ref{eq:tr-tree}. Owing to the token reuse mechanism in tree-based sampling, this ratio is generally less than 1, whereas for parallel sampling it remains fixed at 1.

\begin{align}
\label{eq:tr-tree}
TR_{\text{tree}} = \frac{M*Len*(1+B*\sum_{k=1}^N  k/(N+1)))}{M*Len*(1+B*N)} \\
\text{Simplified:} \quad TR_{\text{tree}} =  \frac{(1+0.5*B*N)}{1+B*N}
\end{align}

Under a common parameter configuration ($N = 3, B = 2$), tree-based sampling requires only 60\% of the tokens consumed by parallel sampling to achieve the same number of rollouts.

Inspired by the entropy-fork tree-structured rollout method, we incorporate it into test-time reinforcement learning, thereby introducing the Entropy-fork Tree-structured Reinforcement Learning (ETRL) framework. This method effectively balances exploration and exploitation at the token level while efficiently reusing low-entropy tokens, thus mitigating the high token consumption issue inherent in test-time reinforcement learning. Experimental results demonstrate that the proposed approach outperforms baseline methods in both efficiency and accuracy, with comprehensive validation detailed in the following sections.

\begin{takeaways}
We propose an Entropy-fork Tree-structured Reinforcement Learning (ETRL) method. During sampling, this approach forks new sampling chains from high-entropy tokens while reusing low-entropy tokens, thereby achieving a token-level balance between exploration and exploitation. Mathematically, the average token consumption of ETRL is expressed as $(1+0.5*B*N)/(1+B*N)$ relative to that of fully parallel solutions. This method effectively mitigates the excessive token cost in existing unsupervised reinforcement learning paradigms while improving estimation accuracy, thereby providing enhanced scalability for large-scale test-time reinforcement learning.
\end{takeaways}

\subsection{EAR: the exploration and exploitation  of Reward learning}
During TTRL training, the policy model generates pseudo-labels via majority voting over sampled responses. In the initial phase, however, the majority ratio is often extremely low (e.g., below 10\% on AIME), meaning that only a small portion of samples obtain positive rewards. After normalization within each rollout group, these few ``lucky'' samples are assigned disproportionately large advantages, which in turn amplify their gradients. 

In supervised reinforcement learning with ground-truth labels, this mechanism facilitates significant convergence toward correct answers. However, in unsupervised scenarios, over-reliance on estimated answers introduces considerable uncertainty. Specifically, in the early stages of training, low estimation accuracy leads the model to assign excessive confidence to incorrect answers, resulting in what is known as premature ``overconfidence''.

\begin{figure}[htbp]
    \centering
    \begin{minipage}{0.32\textwidth}
        \centering
        \includegraphics[width=\linewidth]{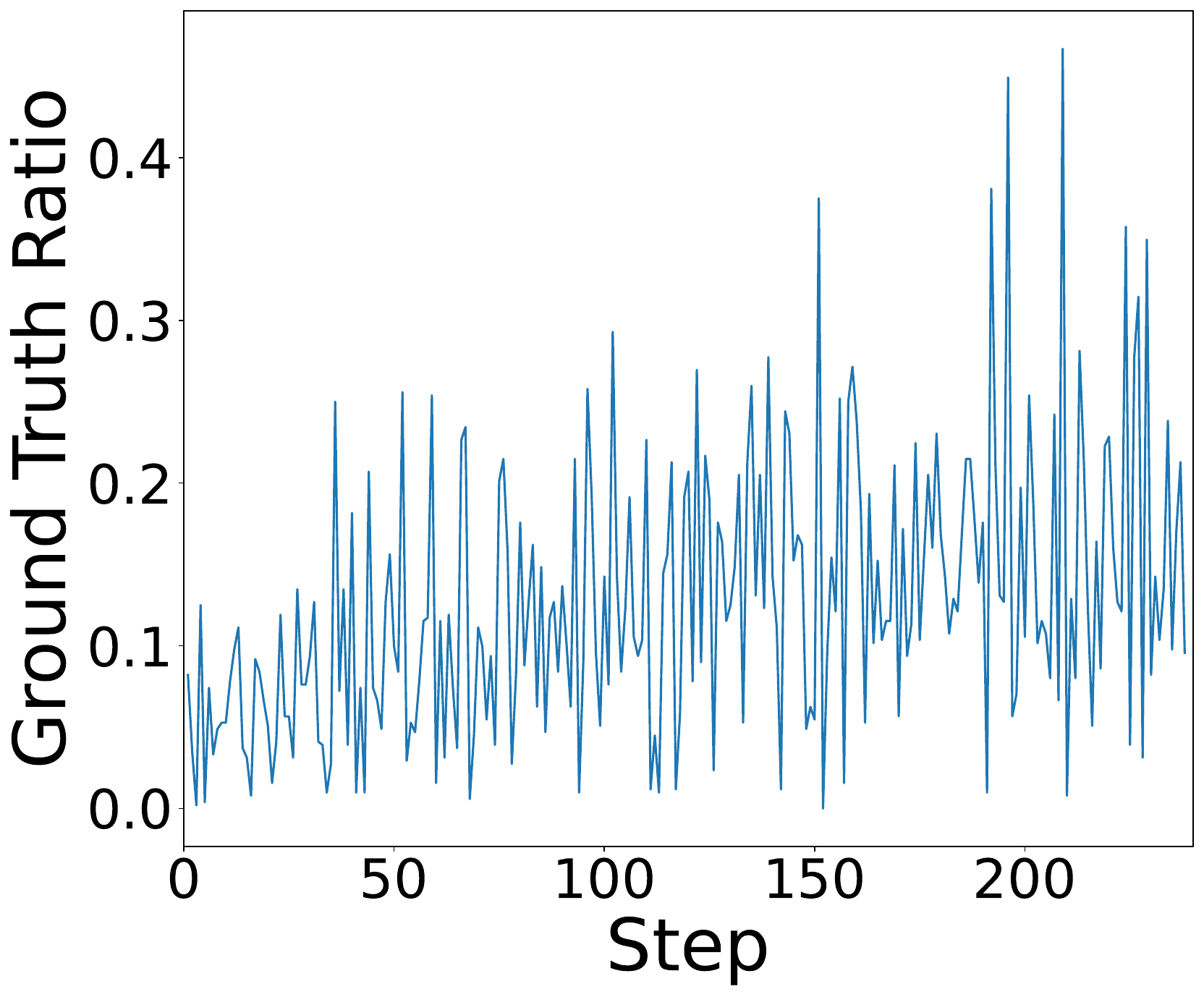}
        \label{fig:img1}
    \end{minipage}
    \begin{minipage}{0.32\textwidth}
        \centering
        \includegraphics[width=\linewidth]{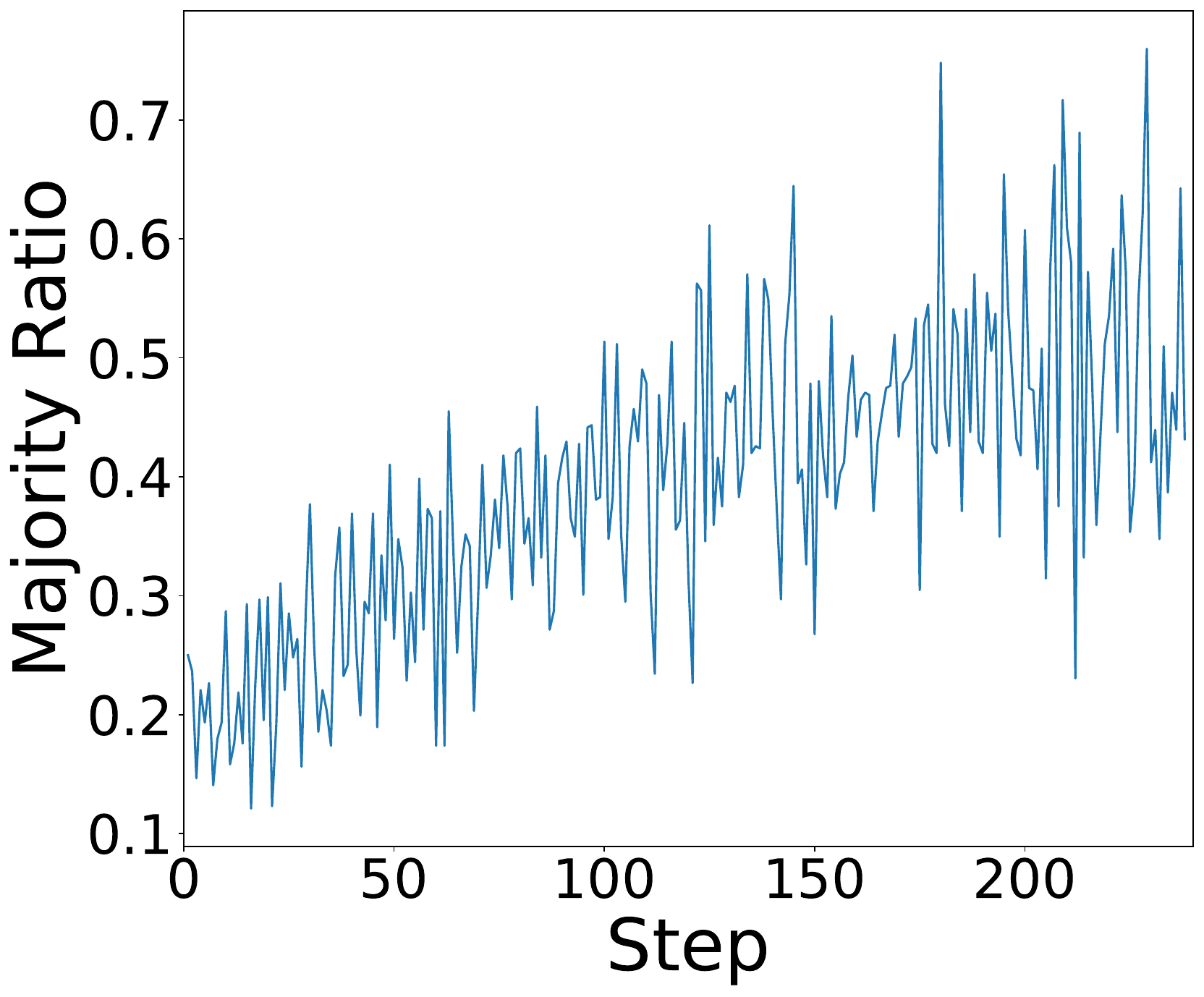}
        \label{fig:img2}
    \end{minipage}
    \begin{minipage}{0.32\textwidth}
        \centering
        \includegraphics[width=\linewidth]{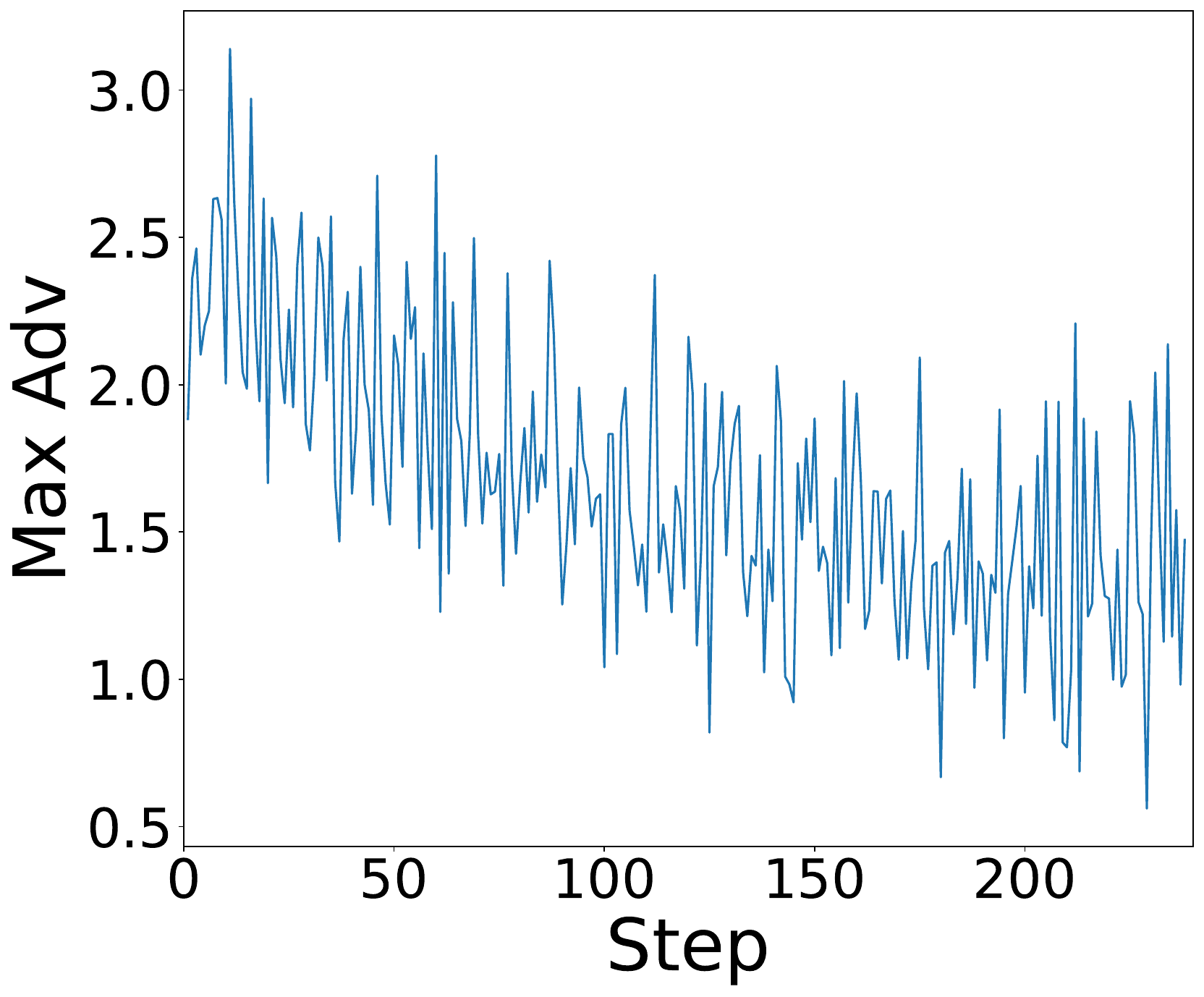}
        \label{fig:img3}
    \end{minipage}
    \caption{During TTRL training on AIME task, the majority ratio progressively increases(middle figure), while the relative advantage among positive sample groups gradually decreases (right figure). However, in the early phase, higher entropy leads to reduced accuracy in majority voting (left figure). Consequently, during the initial stages of consensus-based voting training, lower prediction accuracy paradoxically confers a significant advantage, thereby becoming the source of the model’s overconfidence.}
    \label{fig:three_horizontal_over_confi}
\end{figure}

As illustrated in Figure \ref{fig:three_horizontal_over_confi}, the majority ratio gradually increases from 10\% to 70\%. The figure reveals an exponential negative correlation between the majority ratio and the corresponding reward advantages. During the initial training phase, these low-confidence yet biased advantage signals can easily trap the model in local optima, ultimately leading to suboptimal convergence.

To counteract this instability, we adopt Adv-Clip as the primary regularization strategy. The core idea of Adv-Clip is straightforward yet highly effective: it constrains the magnitude of the advantage values within a predefined range, thereby directly suppressing extreme updates in the early stages of training. Formally, the clipped advantage is expressed as:

\begin{equation}
\label{eq:clip_dav2}
\hat{A}^{clip}_{i,t}  = \text{clip}\bigl(\hat{A}_{i,t} ,-\beta ,+\beta \bigr) \\
\end{equation}

By bounding the scale of policy gradients, Adv-Clip prevents a small number of noisy or low-confidence samples from dominating optimization. This mechanism is particularly crucial in the early phase, when pseudo-label accuracy is low and unstable. Empirically, we observe that clipping stabilizes learning curves, reduces the risk of divergence, and maintains sufficient exploration capacity for later stages. Conceptually, Adv-Clip acts as a safeguard, ensuring that the model does not prematurely collapse its exploration due to overconfident yet unreliable reward signals.

While clipping effectively mitigates overconfidence, it does not exploit finer-grained information about the reliability of each response. 
\citet{cui2025entropy} investigated the impact of entropy mechanisms on reinforcement learning, noting that response entropy can serve as a metric for assessing a model's confidence in its outputs.  

To further refine advantage estimation, we introduce an entropy-based mechanism Adv-Res as a complementary strategy, which is expressed as: 
\begin{align}
\label{eq:e_nrom}
\hat{A}^{res}_{i,t} = Y_{i}* \hat{A}_{i,t}  \\
Y_{i} = 1+(\text{avg}(H_{resp}(o_i))-H_{resp}(o_i))/\text{avg}(H_{resp}(o_i)) \\
\text{avg}(H_{resp}(o_i)) = \frac{1}{G} * \sum_{i=1}^G H_{resp}(o_i)  
\end{align}

Here, $H_{resp}$ denotes the response entropy (defined in Equation \ref{eq:response-entropy-detailed}), and $G$ represents the number of rollouts. 
Adv-Res leverages response entropy to assess relative confidence: responses with higher-than-average entropy are considered uncertain, and their advantages are down-weighted, while low-entropy responses receive slightly amplified updates. This soft adjustment enriches the exploration–exploitation balance and yields additional improvements in performance.

\begin{takeaways}
To address the overestimation bias that inflates advantage estimates in test-time reinforcement learning, we propose an advantage shaping mechanism based on relative-entropy regularization. As a result, it effectively mitigates overconfident value approximations while preserving the directionality of policy improvement.
\end{takeaways}

\section{Experiment}

To systematically evaluate the universality of the proposed method, we select representative models spanning diverse architectural families and parameter scales, including Qwen2.5-Math-1.5B, Qwen2.5-3B, and Llama-3.1-8B. Model performance is evaluated on three canonical mathematical reasoning benchmarks: AIME 2024 \citep{li2024numinamath}, AMC \citep{li2024numinamath}, and MATH-500 \citep{hendrycks2021measuring}. All evaluation protocols and general hyperparameter configurations strictly follow those prescribed in TTRL \citep{zuo2025ttrl}.

\paragraph{Evaluation Metric}
We report pass@1 as the primary evaluation metric. To ensure consistency with prior work, all experiments use greedy decoding for pass@1 computation.

\paragraph{Hyperparameter Configuration}
Training uses a cosine learning rate schedule with a peak value of 5e-7 and the AdamW optimizer to update the policy. During rollout, 64 responses are sampled per prompt at a temperature of 0.6 to facilitate voting-based label estimation and are subsequently downsampled to 32 responses per prompt for training. This vote-then-sample strategy has been empirically validated to reduce computational cost without compromising performance. The maximum generation length is capped at 3072 tokens. The number of training episodes is set to 10, 30, and 80 for MATH-500, AMC, and AIME 2024, respectively, proportional to dataset size. All experiments are conducted on eight NVIDIA A800 80 GB GPUs.


\begin{table}[ht]
\caption{Performance Comparison Between ETMR and TTRL in similar number of rollouts}
\label{tab:etmr_performance}
\begin{center}
{\renewcommand{\arraystretch}{1.2}
\begin{tabular}{lccccc}
\toprule
Model                              & Name      & AIME 2024  & AMC   & MATH-500  & Avg   \\ \midrule
\multirow{3}{*}{Qwen2.5-Math-1.5B} & TTRL      & 15.8       & 48.9  & 73.0      & 45.9  \\
                                   & ETMR      & 21.0       & 50.8  & 76.9      & 49.6  \\
                                   & $\Delta$  & $\uparrow$32.9\%  & $\uparrow$3.9\%  & $\uparrow$5.3\%  & $\uparrow$8.1\% \\ \midrule
\multirow{3}{*}{Qwen2.5-Base-3B}   & TTRL      & 7.9        & 40.7  & 72.2      & 40.3  \\
                                   & ETMR      & 9.2        & 41.7  & 71.7      & 40.9  \\
                                   & $\Delta$  & $\uparrow$16.5\%  & $\uparrow$2.5\%  & $\downarrow$0.7\%  & $\uparrow$1.5\% \\ \midrule
\multirow{3}{*}{Llama-3.1-8B}      & TTRL      & 10.0       & 32.3  & 63.7      & 35.3  \\
                                   & ETMR      & 16.9       & 35.4  & 59.5      & 37.3  \\
                                   & $\Delta$  & $\uparrow$69.0\%  & $\uparrow$9.6\%  & $\downarrow$6.6\%  & $\uparrow$5.7\%    \\
\bottomrule
\end{tabular}
}
\end{center}
\end{table}

\paragraph{Experiment of ETMR}
In the first experiment, we replace TTRL’s fully parallel sampling strategy with our proposed ETMR. Proxy labels are obtained via consensus voting, and subsequent GRPO updates are performed on these labels. For ETMR, we set the hyperparameters as follows: $M$ (number of trees) = 12, $N$ (branching points) = 2, and $B$ (branches per branching point) = 2, yielding an aggregate of 60 rollouts. In contrast, TTRL maintains its original configuration of 64 rollouts --- marginally exceeding ETMR in count. Consistent with the base protocol, both approaches are downsampled to 32 rollouts for gradient computation. Under these settings, equation \ref{eq:tr-tree} shows that ETMR reduces the average token consumption to 60\% of that required by the fully parallel baseline. The performance results are reported in Table \ref{tab:etmr_performance}.

\paragraph{Experiment of EAR}

In the second experiment, we replace the vanilla GRPO advantage estimator with the two advantage-shaping mechanisms described above. For the relative-entropy-scaled advantage (Adv-Res), the scaling function is symmetrically clipped at at $\pm 0.2$; for direct advantage clipping (Adv-Clip), the bounds were set to $\pm 2$. These clipping parameters remain constant across all models and datasets. The performance results are reported in Table \ref{tab:sharping_performance}, and the pass@1 accuracy training curves are shown in Figure \ref{fig:sharping_pic}.

\begin{table}[ht]
\caption{Performance Comparison Between two advantage shaping methods and TTRL}
\label{tab:sharping_performance}
\begin{center}
{\renewcommand{\arraystretch}{1.1}
\begin{tabular}{lccccc}
\toprule
Model                              & Name      & AIME 2024  & AMC   & MATH-500  & Avg  \\ \midrule
\multirow{4}{*}{Qwen2.5-Math-1.5B} & TTRL      & 15.8       & 48.9  & 73.0      & 45.9    \\
                                   & Adv-Res   & 19.6       & 51.0  & 77.3      & 49.3    \\
                                   & Adv-Clip  & 19.4       & 50.5  & 77.3      & 49.1    \\ 
                                   & $\Delta$  & $\uparrow$24.1\%  & $\uparrow$4.3\%  & $\uparrow$5.9\%  & $\uparrow$7.4\% \\ \midrule
\multirow{4}{*}{Qwen2.5-Base-3B}   & TTRL      & 7.9        & 40.7  & 72.2      & 40.3    \\
                                   & Adv-Res   & 13.1       & 41.4  & 72.4      & 42.3    \\
                                   & Adv-Clip  & 10.0       & 42.0  & 71.3      & 41.1    \\ 
                                   & $\Delta$  & $\uparrow$65.8\%  & $\uparrow$3.2\%  & $\uparrow$0.3\%  & $\uparrow$5.0\% \\ \midrule
\multirow{4}{*}{Llama-3.1-8B}      & TTRL      & 10.0       & 32.3  & 63.7      & 35.3    \\
                                   & Adv-Res   & 13.5       & 36.4  & 61.3      & 37.1    \\
                                   & Adv-Clip  & 13.5       & 34.7  & 63.2      & 37.1    \\ 
                                   & $\Delta$  & $\uparrow$35.0\%  & $\uparrow$12.7\%  & $\downarrow$0.8\%  & $\uparrow$5.1\% \\ 
\bottomrule
\end{tabular}
}
\end{center}
\end{table}

As shown in Table \ref{tab:sharping_performance}, both the advantage-scaling and advantage-clipping variants yield consistent gains over the native GRPO advantage estimator across datasets and model scales. For example, on the AIME 2024 benchmark, Adv-Res increases the Qwen2.5-3B pass@1 by 65\% over the baseline. The improvements are less pronounced for specialized mathematical models and larger architectures, which we attribute to their lower epistemic uncertainty. By contrast, smaller, non-mathematical models exhibit higher uncertainty on reasoning-intensive tasks, making them more susceptible to overconfident value estimates.

When directly comparing the two regularization strategies, relative-entropy-scaled advantage shaping (Adv-Res) consistently outperforms direct clipping (Adv-Clip). By softly penalizing high-entropy outputs while encouraging cautious exploration in low-entropy regions, Adv-Res achieves a more stable balance between exploitation and exploration.

\begin{takeaways}
The Entropy-fork Tree Majority Rollout (ETMR) method demonstrates superior efficiency and effectiveness in consistent estimation reinforcement learning, exhibiting an average token consumption of merely 60\% compared to fully parallel approaches. This provides feasibility support for scaling large-scale unsupervised reinforcement learning in subsequent research.
The advantage-shaping mechanism significantly enhances mathematical reasoning performance in unsupervised reinforcement learning, with particularly pronounced effects observed in smaller models trained on non-mathematical instructions.
\end{takeaways}

\section{Discussions}

\subsection{Why is the ETMR method effective?}

The efficiency of ETMR has been demonstrated in the preceding sections, accompanied by a mathematical derivation of its average efficiency improvement. Experimental results show that for more challenging datasets (e.g., AIME), ETMR yields greater relative improvements compared to easier datasets. Our previously proposed hypothesis suggests that ETMR branches on high-entropy tokens, thereby exhibiting stronger inherent exploratory capabilities than fully parallel strategies. This mechanism enables proxy labels to achieve higher accuracy. 
ETMR demonstrates significant improvements over baseline methods on non-Math models and challenging datasets, thereby enhancing the overall precision of subsequent policy models. We find that proxy label accuracy directly influences final performance, a result that both supports and validates our hypothesis.

Furthermore, we attempted to enhance diversity by adjusting the temperature coefficient. 
Tests on the base model revealed that excessively high temperature coefficients degrade overall performance. Directly increasing the temperature coefficient significantly reduces the accuracy of proxy labels obtained through consensus voting and decreases the initial majority ratio. This may cause an imbalance in group reward distribution and lead to model overconfidence issues.

\subsection{Why use relative entropy to shape advantages instead of absolute entropy?}

Initially, following prior work, we adopted absolute entropy as the shaping basis. However, we identified a dimensional inconsistency between advantages and entropy, and noted that absolute entropy is influenced by multiple factors. Although we validated the effectiveness of absolute entropy shaping across multiple datasets and models, its implementation demands extensive hyperparameter tuning. To improve the method’s generalizability, we shifted to the concept of relative entropy. Additionally, in our experiments, we compared this approach with a simple advantage clipping method, further validating the effectiveness of relative entropy.

\section{Limitation}

Although ETMR offers a theoretical reduction in token consumption, the observed wall-clock acceleration falls short of the theoretical expectation. This discrepancy stems from the differing utilization characteristics of the two sampling paradigms: fully parallel sampling exploits batched execution to saturate GPU capacity, whereas ETMR relies on a tree-structured, pipeline-style rollout. The current RL training framework (Verl) lacks native support for such hybrid execution patterns; extending Verl’s scheduling primitives is left for future work. Empirically, ETMR also exhibits pronounced sensitivity to the temperature parameter --- excessively high values precipitate training collapse. Likewise, both the relative-entropy scaling coefficient and the clipping bounds substantially affect final accuracy, and their optimal values appear to be dataset- and model-dependent. A principled search over these hyperparameters is beyond the scope of the present study.

\bibliography{iclr2026_conference}
\bibliographystyle{iclr2026_conference}

\newpage
\appendix
\section{Appendix}
1

\begin{figure}[ht]
\centering
\begin{subfigure}{0.32\textwidth}
    \includegraphics[width=\linewidth]{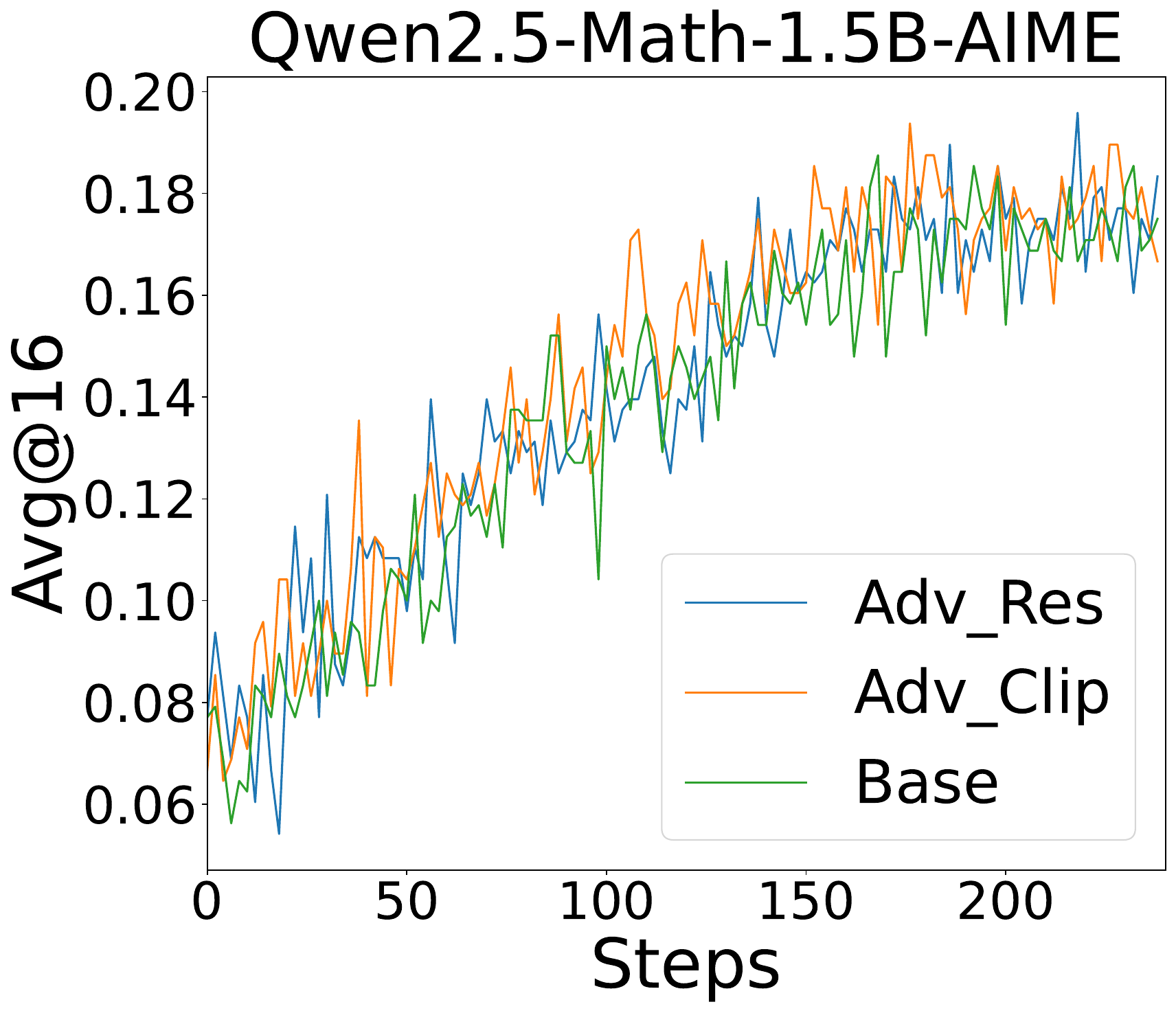}
    \caption{AIME24 scores of Qwen2.5-Math-1.5B.}
\end{subfigure}
\begin{subfigure}{0.32\textwidth}
    \includegraphics[width=\linewidth]{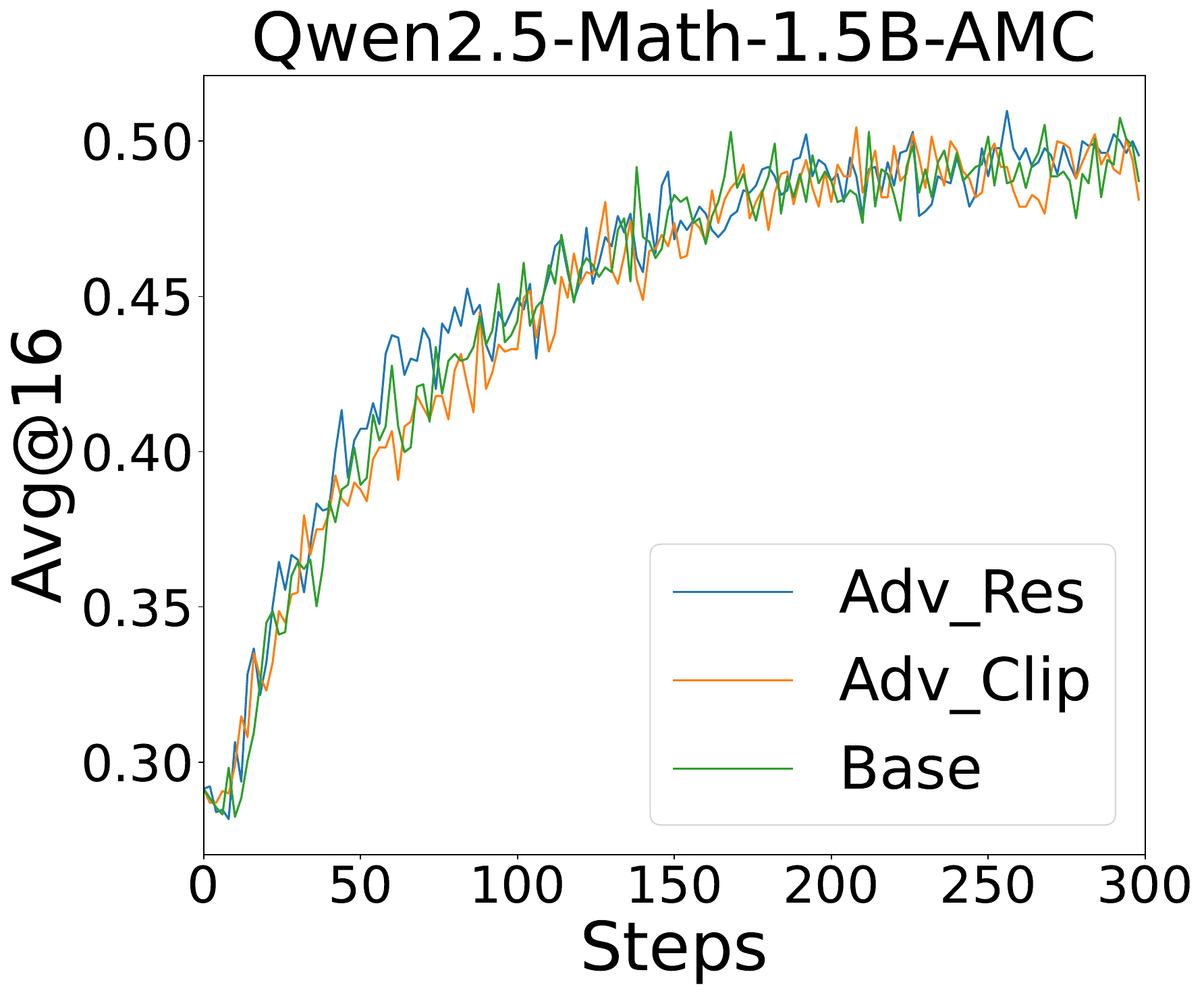}
    \caption{AMC scores of Qwen2.5-Math-1.5B.}
\end{subfigure}
\begin{subfigure}{0.32\textwidth}
    \includegraphics[width=\linewidth]{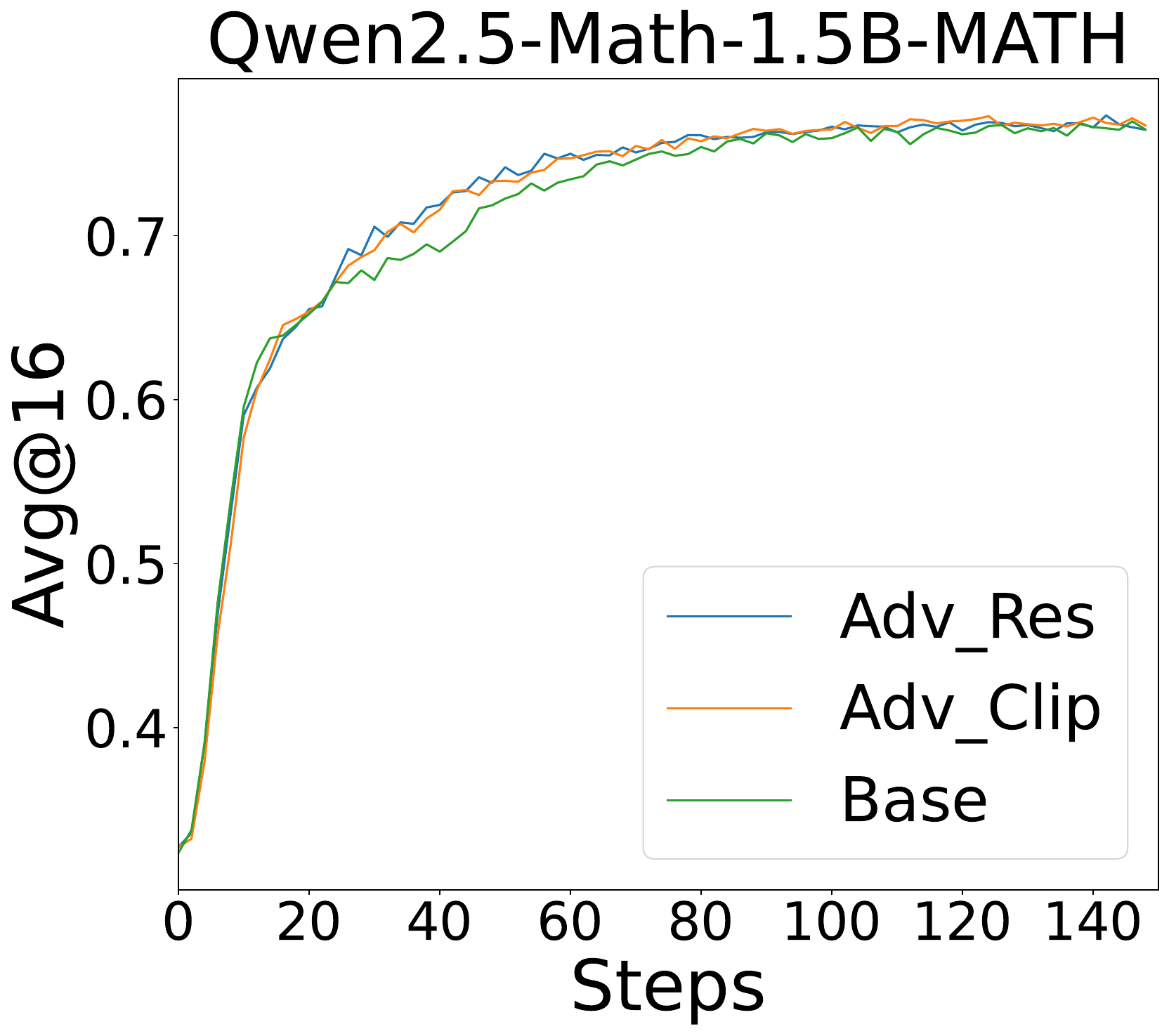}
    \caption{MATH scores of Qwen2.5-Math-1.5B.}
\end{subfigure}

\begin{subfigure}{0.32\textwidth}
    \includegraphics[width=\linewidth]{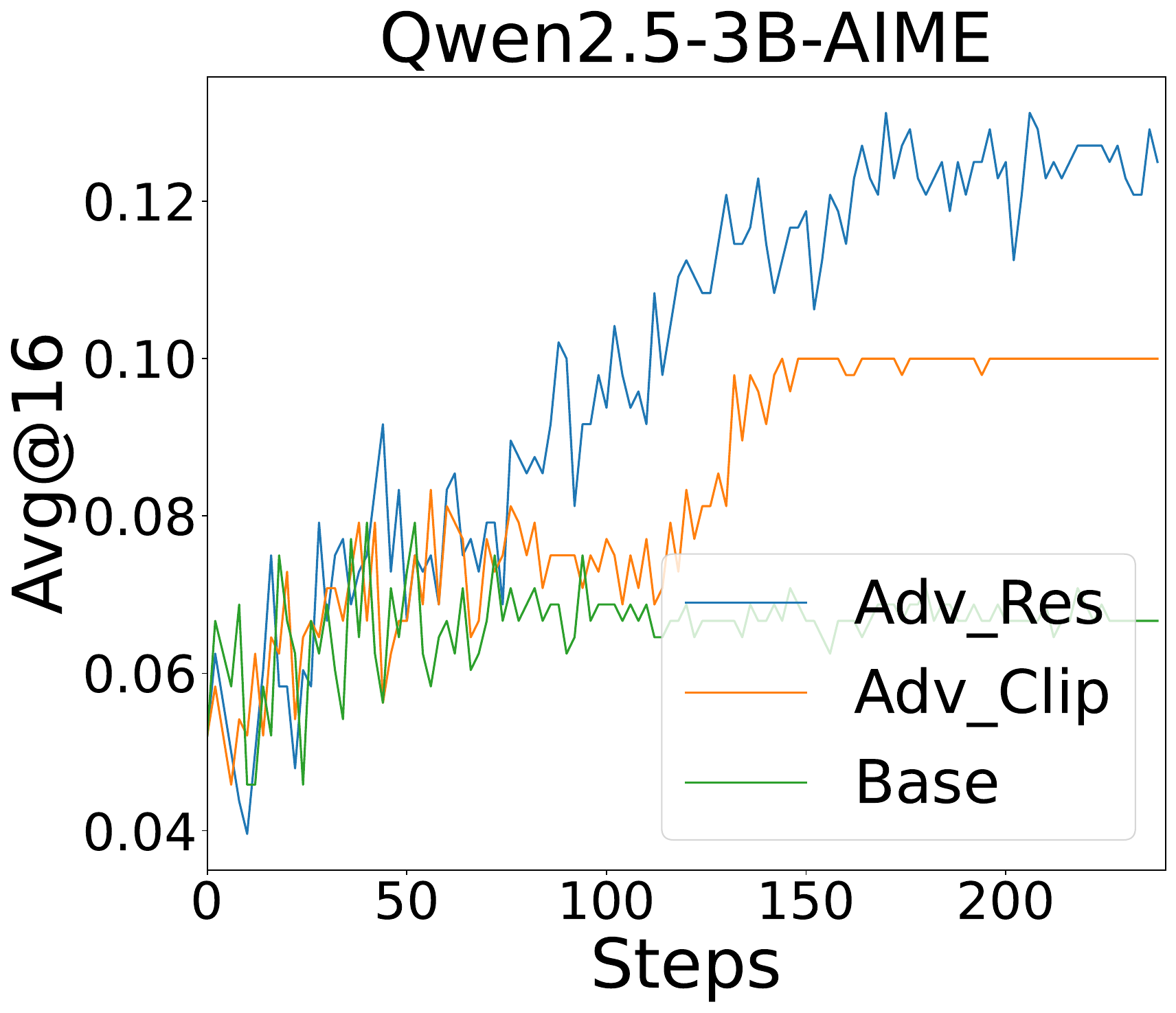}
    \caption{AIME24 scores of Qwen2.5-Base-3B.}
\end{subfigure}
\begin{subfigure}{0.32\textwidth}
    \includegraphics[width=\linewidth]{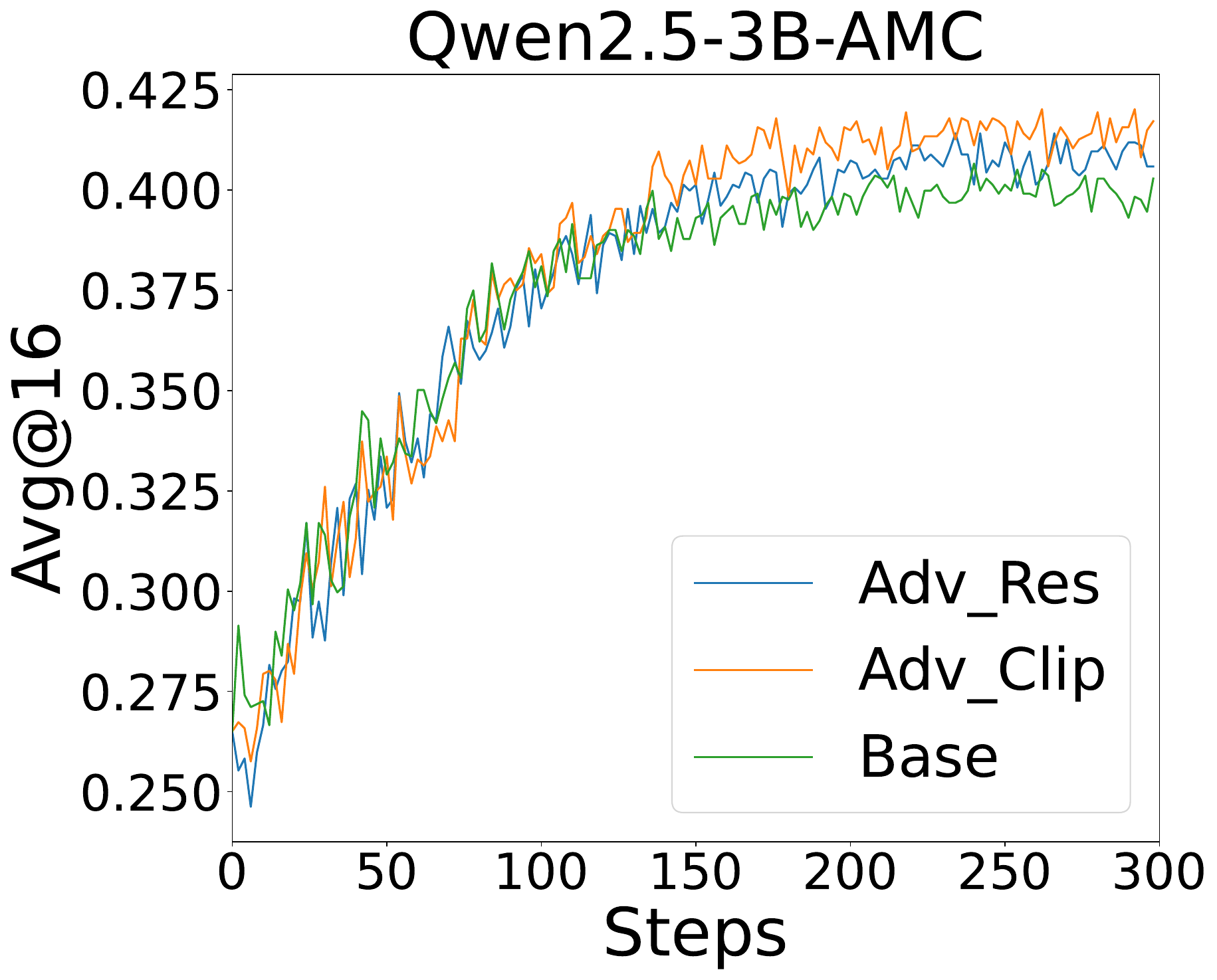}
    \caption{AMC scores of Qwen2.5-Base-3B.}
\end{subfigure}
\begin{subfigure}{0.32\textwidth}
    \includegraphics[width=\linewidth]{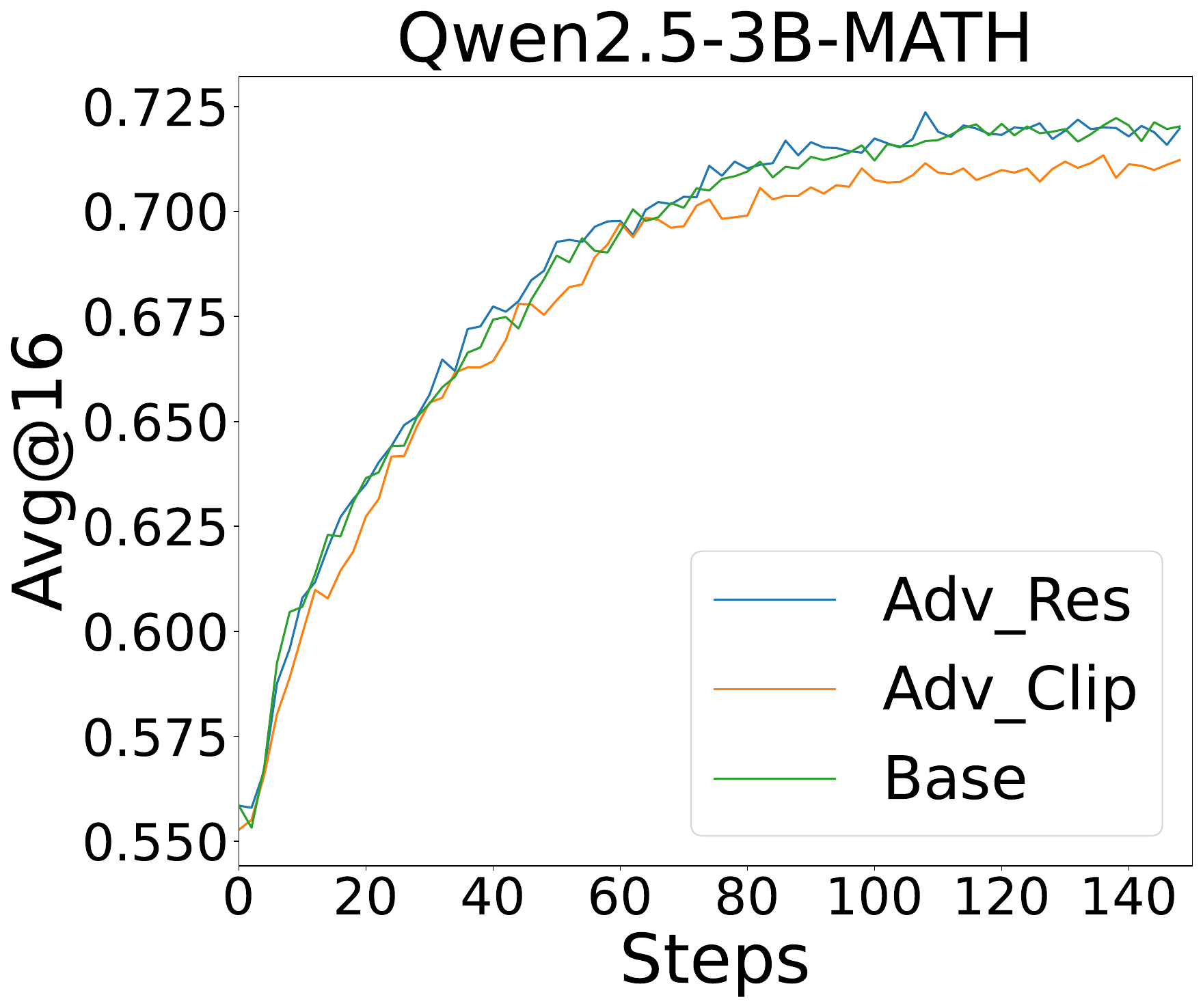}
    \caption{MATH scores of Qwen2.5-Base-3B.}
\end{subfigure}

\begin{subfigure}{0.32\textwidth}
    \includegraphics[width=\linewidth]{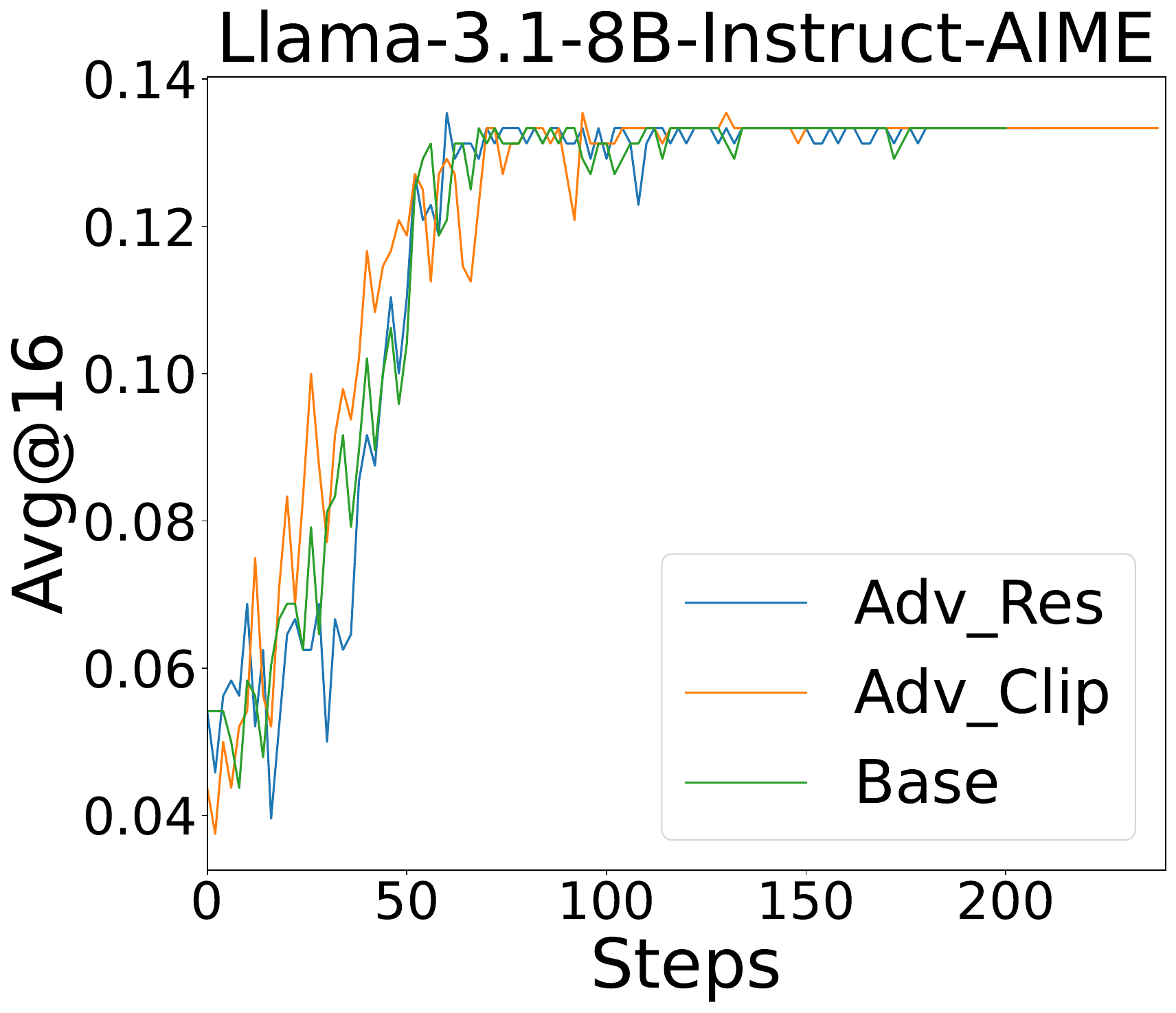}
    \caption{AIME24 scores of Llama-3.1-8B.}
\end{subfigure}
\begin{subfigure}{0.32\textwidth}
    \includegraphics[width=\linewidth]{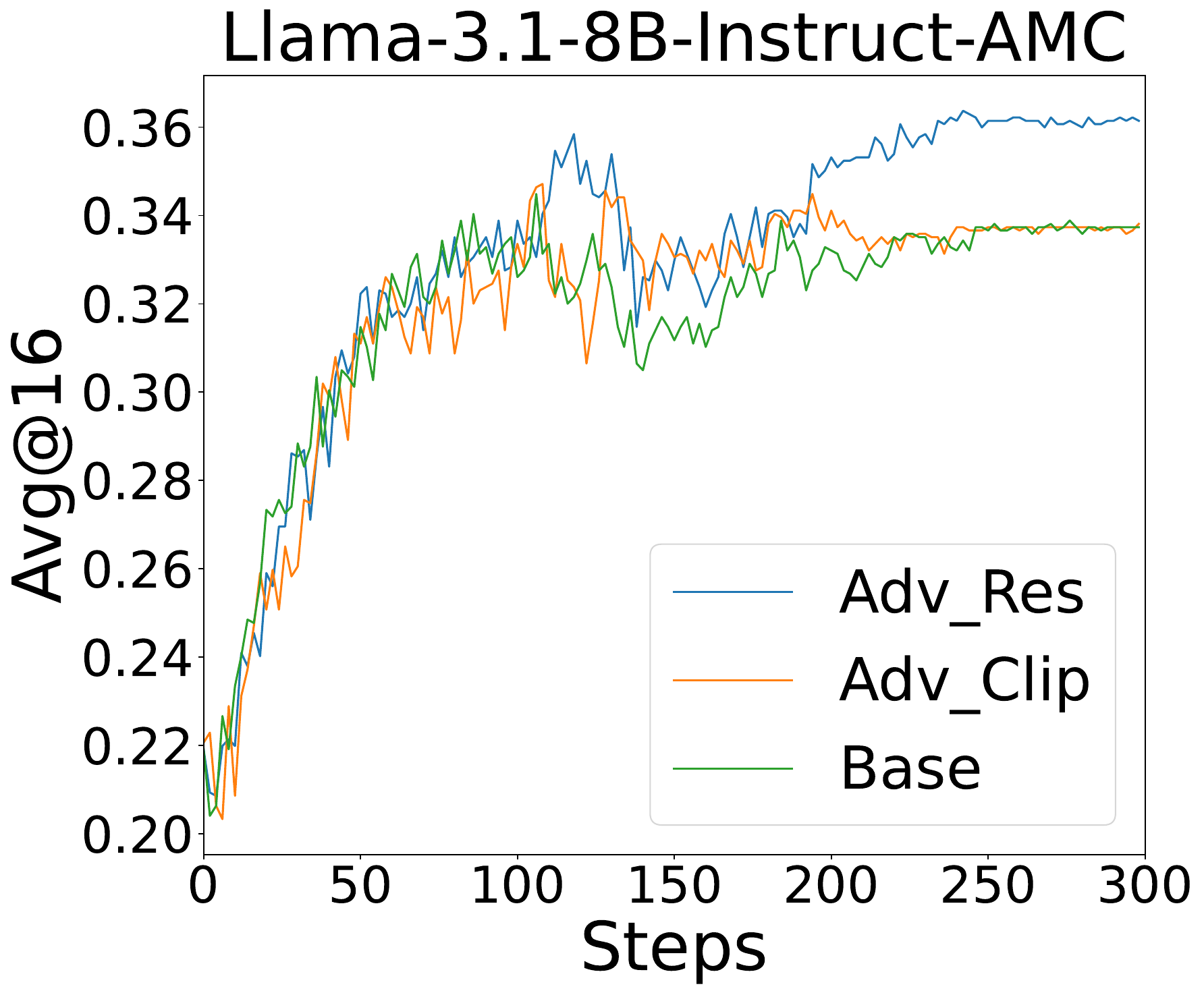}
    \caption{AMC scores of Llama-3.1-8B.}
\end{subfigure}
\begin{subfigure}{0.32\textwidth}
    \includegraphics[width=\linewidth]{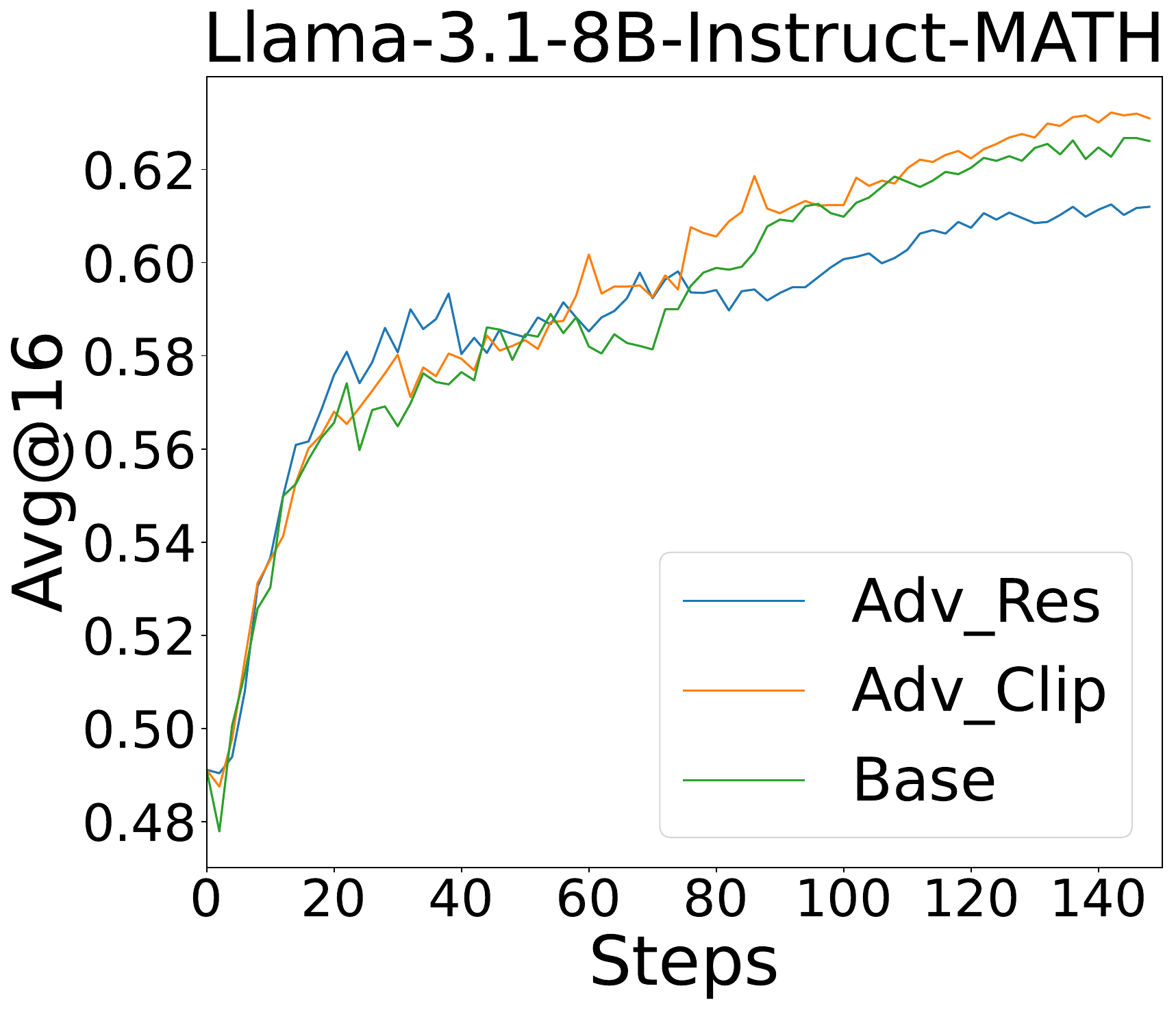}
    \caption{MATH scores of Llama-3.1-8B.}
\end{subfigure}

\caption{Comparison of between ADV-RES and ADV-CLIP}
\label{fig:sharping_pic}
\end{figure}

\end{document}